\documentclass{article}

\usepackage{microtype}
\usepackage{graphicx}
\usepackage{subfigure}
\usepackage{booktabs} % for professional tables

\usepackage{hyperref}

\usepackage[accepted]{icml2021}

\icmltitlerunning{A Dual Process Model for Optimizing Cross Entropy in Neural Networks}

\begin{document}

\twocolumn[
\icmltitle{A Dual Process Model for Optimizing Cross Entropy in Neural Networks}



\icmlsetsymbol{equal}{*}

\begin{icmlauthorlist}
\icmlauthor{Stefan Jaeger}{nih}
\end{icmlauthorlist}

\icmlaffiliation{nih}{National Library of Medicine, National Institutes of Health, Bethesda, MD, USA}

\icmlcorrespondingauthor{Stefan Jaeger}{\mbox{stefan.jaeger@nih.gov}}

\icmlkeywords{Machine Learning, Neural Networks, Optimization, Cross Entropy, Backpropagation, Golden Ratio}

\vskip 0.3in
]



\printAffiliationsAndNotice{}  % leave blank if no need to mention equal contribution

%%%%%%%%%%%%%%%%%%%%%%%%%%%%%%%%%%%%%%%%%%%%%%
% Main Text
%%%%%%%%%%%%%%%%%%%%%%%%%%%%%%%%%%%%%%%%%%%%%%

\begin{abstract}
Minimizing cross-entropy is a widely used method for training artificial neural networks. Many training procedures based on backpropagation use cross-entropy directly as their loss function. Instead, this theoretical essay investigates a dual process model with two processes, in which one process minimizes the Kullback–Leibler divergence while its dual counterpart minimizes the Shannon entropy. Postulating that learning consists of two dual processes complementing each other, the model defines an equilibrium state for both processes in which the loss function assumes its minimum. An advantage of the proposed model is that it allows deriving the optimal learning rate and momentum weight to update network weights for backpropagation. Furthermore, the model introduces the golden ratio and complex numbers as important new concepts in machine learning.
\end{abstract}

\section{Introduction}
\label{Sec:Introduction}

Modern neural networks are typically trained on big data in a supervised fashion. This is commonly achieved by minimizing a loss function that measures the distance between network output and teaching input in several iterations so that the network output approaches the training input. A widely used method to do this is backpropagation, which descends along the gradient of the loss function by propagating weight changes through the network~\cite{rumelhart1986learning,bengio2012practical,lecun2012efficient}. While backpropagation has been applied very successfully, there are still several open questions requiring a definite answer. In particular, the type of loss function to be used and the optimal way of following the gradient are still open problems. A variety of loss functions has been used in the literature to judge the quality of a network output. Similarly, many optimization strategies have been investigated to let gradient descent converge faster or avoid local minima, including stochastic gradient descent, second order methods, and others~\cite{bengio2012practical,sutskever2013importance}. However, most design decisions are still largely based on empirical experiments and experience.

This paper presents a theoretical study of one of the most popular loss functions, cross-entropy. Technically, the paper is an advancement of an approach developed in~\cite{jaeger2013neurological, jaeger2020arXiv}. Specifically, it discusses the theoretical ramifications when cross-entropy is minimized by two dual processes, which minimize the Kullback–Leibler divergence and the Shannon entropy, respectively. The advantage of this approach is that it leads to a loss function and weight space for which learning rate and momentum weight can be derived theoretically. These regularization parameters are important parameters for weight updates during backpropagation and thus have a strong influence on gradient decent. Although rules of thumb and effective heuristics exist for choosing these parameters, including dynamic parameter updates during backpropagation, they have eluded a conclusive theoretical analysis so far. The theoretical values derived in the following are similar to empirical values in the literature. Another advantage of the dual process model proposed in this paper is that it includes the golden ratio and complex numbers, which are both novel concepts in machine learning. Complex numbers in particular seem to be largely ignored by the vast majority of publications in computational machine learning. However, incorporating complex numbers may be necessary to achieve a full understanding of neural networks, in view of brain waves or similarities with physical systems, for example. The model proposed in this paper is an attempt at doing so.

The paper is structured as follows: Section~\ref{Sec:MotivationApproach} motivates the paper, discussing intrinsic uncertainty in human perception. Section~\ref{Sec:CrossEntropy} recalls the well-known definitions of cross-entropy, Shannon entropy, and Kullback–Leibler divergence. Then, Section~\ref{Sec:GoldenRatio} lists the mathematical features of the golden ratio, borrowing text from an earlier work~\cite{jaeger2020arXiv}, before Section~\ref{Sec:IntrinsicUncertainty} formalizes the intrinsic uncertainty between observed and true probabilities. Based on these results, Section~\ref{sec:LossFunction} will derive a new loss function. Next, Section~\ref{Sec:ProcessModel} will present the dual process model from which Section~\ref{Sec:Regularization} will then derive the two regularization parameters, learning rate and momentum weight. Finally, a discussion and conclusion will summarize the results of the paper.

\section{Motivation and Approach}
\label{Sec:MotivationApproach}

\begin{figure}[ht] % h-here, t-top, (b-bottom)
\vskip 0.2in
\begin{center}
%\centerline{\includegraphics[width=\columnwidth]{figures/Rubin2.jpg}}
\centerline{\includegraphics[width=\columnwidth]{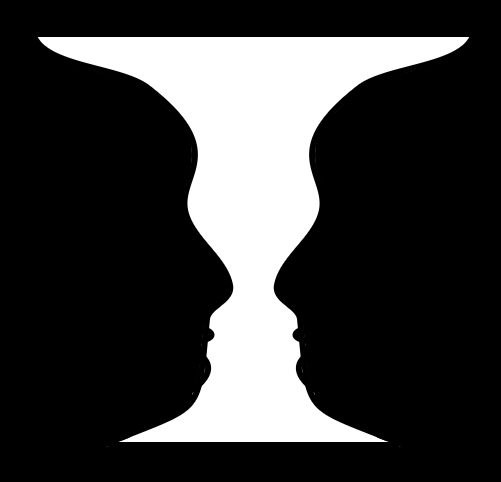}}
\caption{Rubin's Vase}
\label{RubinVaseFigure}
\end{center}
\vskip -0.2in
\end{figure}

The main motivation of the approach developed in this paper lies in the observation that human perception is fraught with intrinsic uncertainty. A good example is Rubin's Vase, an ambiguous visual perception made popular by the Danish psychologist Edgar Rubin around 1915, which is shown in Figure~\ref{RubinVaseFigure}~\cite{rubin2015}.
This figure could be interpreted either as a vase, or as two faces looking at each other, depending on what is considered background and foreground. One interpretation is the complement of the respective other. To develop this idea further in a more formal approach, let $\mathcal{X}$ be a discrete random variable with two possible values, $\mathcal{X} = \left\{x, \neg x \right\}$. Furthermore, let $\mathrm{P} (\mathcal{X})$ be a probability mass function on $\mathcal{X}$  that assigns probability values as follows: $\mathrm{P}(x) = p$, and $\mathrm{P}(\neg x) = 1-p$. Assuming that $\mathcal{X}$ underlies the intrinsic uncertainty of our perception, viewers do not know whether they observe $x$ or $\neg x$, or in information-theoretical terms, they do not know whether the information content of their observation is $-\ln(p)$ or $-\ln(1-p)$. Assuming, without loss of generality, that $p$ is the true probability, viewers do not know whether the information they can expect is $-p \cdot \ln(p)$ or $-p \cdot \ln(1-p)$. Only if both terms are equal, which is the case for $p=0.5$ when $p \in \; ]0\,;1[\,$, is there no uncertainty between them:
\begin{equation}
-p \cdot \ln(p) = -p \cdot \ln(1-p),
\label{motivationEquation1}
\end{equation}
or equivalently:
\begin{equation}
0 = -p \cdot \ln\bigg(\frac{1-p}{p}\bigg) 
\label{motivationEquation2}
\end{equation}
In Eq.~\ref{motivationEquation2}, the intrinsic uncertainty shows as follows: If an observer knows the value of~$p$, the observer does not know whether the fraction $(1-p)/p$ or its inverse needs to be used as argument to the logarithmic term, and for that matter, does not know the sign of the difference. Conversely, if the observer knows the sign of the difference, the observer cannot know the value of~$p$, which could be either~$p$ or~$1-p$.

Eq.~\ref{motivationEquation2} also shows the basic equation structure of the Kullback–Leibler divergence, which will be discussed in the next section and Section~\ref{Sec:IntrinsicUncertainty}.

\section{Cross Entropy}
\label{Sec:CrossEntropy}

Cross-entropy is a common loss function used for training of artificial neural networks. It quantifies the difference between two probability distributions, say~$p$ and~$q$. In communication theory, it measures the average number of bits needed to encode data coming from a source with distribution~$p$, when the model (encoding) is optimized for an estimated probability distribution~$q$, rather than the true distribution~$p$.

Mathematically, for discrete probability distributions~$p$ and~$q$, defined on the same probability space $\mathcal{X}$, the cross-entropy~$H(p,q)$ is computed as follows:
\begin{equation}
H(p,q) = -\sum_{x \in \mathcal{X}} p(x) \cdot \ln\Big( q(x)\Big)
\label{crossEntropyEquation}
\end{equation}

The cross-entropy~$H(p,q)$ can also be expressed as the sum of the Kullback–Leibler divergence~$D(p,q)$ from~$p$ to~$q$ and the Shannon entropy~$H(p)$:
\begin{equation}
H(p,q) = D(p,q) + H(p)
\label{crossEntropySumEquation}
\end{equation}
The next subsections will briefly write out the definitions for these two information measures, including the special case of a two-valued random variable~$\mathcal{X}$ with outcome probabilities~$p$ and~$1-p$.

\subsection{Kullback–Leibler divergence}

The Kullback–Leibler divergence~$D_{KL}(p,q)$ describes the difference between a probability distribution~$p$, say a measured observation, and a second probability distribution, $q$, serving as a reference or model distribution. The Kullback–Leibler divergence can then be interpreted as the average difference of the number of bits required for encoding samples of~$p$ using the optimal encoding given by~$q$ (rather than the optimal coding for~$p$). 

For the two distributions, $p$ and~$q$, the Kullback–Leibler divergence~$D_{KL}(p,q)$ is then defined as follows
\begin{equation}
D_{KL}(p,q) = -\sum_{x \in \mathcal{X}} p(x) \cdot \ln \Bigg(\frac{q(x)}{p(x)}\Bigg)
\label{KullbackLeiblerEquation}
\end{equation}
for a probability space $\mathcal{X}$.

In the specific case of a two-valued random variable~$p$, the Kullback–Leibler divergence $D_{KL}(p,\neg p)$ between~$p$ and its complement~$1-p$ therefore computes as follows
\begin{equation}
D_{KL}(p,\neg p) = -p \cdot \ln \Bigg(\frac{1-p}{p}\Bigg) - (1-p) \cdot \ln \Bigg(\frac{p}{1-p}\Bigg)
\label{KullbackLeiblerEquation2}
\end{equation}
For this specific case, the divergence $D(p) = D_{KL}(p,\neg p)$, assumes its minimum of zero when both distributions $p$ and $\neg p$ are identical, and thus when both outcomes of the random variable have a probability of~$0.5$.

\subsection{Shannon Entropy}

The Shannon entropy, or simply entropy, of a random variable $\mathcal{X}$ is the average information conveyed by its possible outcomes~\cite{shannon1948mathematical}. In mathematical terms, the entropy $H(\mathcal{X})$ of~$\mathcal{X}$ can be computed as follows:
\begin{equation}
H(\mathcal{X}) = -\sum_{x \in \mathcal{X}} p(x) \cdot \ln\Big(p(x)\Big),
\label{entropyEquation}
\end{equation}
where~$p(x)$ is a probability distribution defined on all outcomes~$x$ of~$\mathcal{X}$.

For a random variable~$\mathcal{X}$ with two possible outcomes, with probabilities~$p$ and $1-p$, the entropy thus computes as follows:
\begin{equation}
H(\mathcal{X}) = -p \cdot \ln(p) - (1-p) \cdot \ln(1-p)
\label{entropyEquation2}
\end{equation}
In this case, contrary to the Kullback–Leibler divergence, the entropy assumes its maximum (not its minimum) when the outcome and its complement have the same probability, namely $p=0.5$.

\section{Golden Ratio}
\label{Sec:GoldenRatio}

This section highlights a connection between Eq.~\ref{motivationEquation2} and the golden ratio, as shown in~\cite{jaeger2020arXiv}. In Eq.~\ref{motivationEquation2}, the argument to the logarithm, $(1-p)/p$, can be regarded as the perceived (or measured) probability, whereas the multiplier~$p$ is the true probability. If an observer wants to measure the true probability, both terms need to be equal, meaning
\begin{equation}
p  = \frac{1-p}{p}
\label{goldenRatioEquation}
\end{equation}
This holds true if~$p$ is the golden ratio~\cite{livioGoldenRatioBook}, as discussed next.

Eq.~\ref{goldenRatioEquation} is equivalent to the following quadratic equation:
\begin{equation}
p^2 + p - 1  = 0,
\label{plusGoldenRatioEquation}
\end{equation}
which has two irrational solutions~$p_1$ and~$p_2$:
\begin{equation}
p_1 = \frac{\sqrt{5}-1}{2} \approx 0.618,
\label{p1Equation}
\end{equation}
and
\begin{equation}
p_2 = \frac{-\sqrt{5}-1}{2} \approx -1.618
\label{p2Equation}
\end{equation}
An important feature of both solutions is that their complement, $1-p$, equals their square,
\begin{equation}
1-p = p^2
\label{complementPlusEquation}
\end{equation}

Alternatively, the golden ratio can be derived from another quadratic equation, which may be more commonly found in textbooks, and which results from replacing $p$ by $-p$ in Eq.~\ref{plusGoldenRatioEquation}:
\begin{equation}
p^2 - p - 1  = 0
\label{minusGoldenRatioEquation}
\end{equation}
This equation also has two irrational solutions, which are the negatives of~$p_1$ and~$p_2$:  
\begin{equation}
-p_1 \approx -0.618 \mbox{\quad and} -p_2 \approx 1.618
\label{minusp1p2Equation}
\end{equation}
However, for both solutions of the second quadratic equation, Eq.~\ref{minusGoldenRatioEquation}, the complement $1-p$ is the negative reciprocal:
\begin{equation}
1-p = -\frac{1}{p}
\label{complementMinusEquation}
\end{equation}

As mentioned in~\cite{jaeger2020arXiv}, the sum of both solutions for Eq.~\ref{plusGoldenRatioEquation} and Eq.~\ref{minusGoldenRatioEquation} is either minus one or one, respectively:
\begin{equation}
p_1 + p_2 = -1 \mbox{\quad and \quad} -p_1 - p_2 = 1
\label{sumOneEquation}
\end{equation}
The literature sometimes uses the letter $\varphi$ for the golden ratio, and typically defines the golden ratio as a single value, often with $\varphi \approx 1.618$ and discarding negative values. In this paper, all four solutions to the quadratic equations above will be referred to as the golden ratio.

\section{Intrinsic Uncertainty}
\label{Sec:IntrinsicUncertainty}

For Eq.~\ref{motivationEquation2}, the previous section showed that the measured probability equals the true probability in the golden ratio. This section develops Eq.~\ref{motivationEquation2} in a way that allows computing all possible measurements in a systematic way~\cite{jaeger2013neurological, jaeger2020arXiv}.
%~\cite{Jaeger2012TCM}

By coupling the measured probability with the true probability, according to the relationship in~Eq.\ref{goldenRatioEquation}, and using the letter~$E$ (Energy) for the information difference, Eq.~\ref{motivationEquation2} can be developed as follows:
\begin{eqnarray}
E & = & -p \cdot \ln\bigg(\frac{1-p}{p}\bigg) \label{computationalModelEquation}\\
& \Leftrightarrow & -p^2 \cdot \ln\Big(1-p\Big) \\
& \Leftrightarrow & -p \cdot \ln\Big(1-p^2\Big) \\
& \Leftrightarrow & -p \cdot \ln\Big(\sqrt{1-p^2}\Big) \cdot 2 \label{squareRootPythagoreanEquation}\\
& \Leftrightarrow & -\sin(\phi) \cdot \ln\big(\cos(\phi)\big) \cdot 2,
\label{pythagoreanEquation}
\end{eqnarray}
where the last expression holds for an angle~$\phi \in \big[0\,;\frac{\pi}{2}\big]$. Varying~$\phi$ in this range will produce all possible measurements, which are points on the unit circle. Using this scheme, the measured probability equals the true probability for $\phi = \pi/4$, and a measured probability of $\sin(\pi/4) = \cos(\pi/4) = 1/\sqrt{2}$. The letter~$E$ for energy is used in Eq.~\ref{computationalModelEquation} to emphasize that information is related to energy in the physical sense and also to emphasize that this energy, or information, can be absorbed or released, depending on the sign.

In Eq.~\ref{pythagoreanEquation}, the energy~$E$ attains its minimum of zero for $\phi=0$, when $\sin(\phi)=0$ and $\cos(\phi)=1$. On the other hand, $E$ reaches its maximum, infinity, for $\phi=\pi/2$, when $\sin(\phi)=1$ and $\cos(\phi)=0$. 

Now, a dual energy for a second observer can be computed by swapping $\sin$ for $\cos$ in Eq.~\ref{pythagoreanEquation}, which amounts to using the main diagonal as a mirror axis. This dual energy will reach its maximum, when the original energy uses its minimum; and vice versa, it will be minimum when the original energy is maximum. Most importantly, it is possible to establish a formal analogy to the intrinsic uncertainty in observations, as motivated in Section~\ref{Sec:MotivationApproach}. For this, let there be two dual and intertwined processes based on the dual energies above. While one process considers all energies for $\phi > \pi/4$ as released, and all energies for $\phi < \pi/4$ as absorbed; its dual counterpart considers energies for $\phi > \pi/4$ as absorbed, and all energies for $\phi < \pi/4$ as released. The intrinsic uncertainty then shows as follows: If one process knows whether energy is released or absorbed, it does not know the magnitude of the energy. Conversely, if a process knows the magnitude of the energy, it does not know whether the energy is released or absorbed. Only when both processes work hand in hand, synchronously, can there be knowledge about the direction and magnitude of energy. However, an observer can truly measure only one or the other, which shows a connection to Heisenberg's uncertainty principle~\cite{jaeger2013neurological}.

Nevertheless, an observer can determine the degree of uncertainty in a measurement by varying the angle $\phi$. For example, the observer could choose an angle of $\phi = \pi/4$ for which the observer may know the magnitude of the energy but not its direction, and then vary the angle to learn more about the direction at the expense of knowing about magnitude. Mathematically, this regularization can be achieved by a multiplier that regulates the energy~$E$, as shown in the following modification of Eq.~\ref{computationalModelEquation}:
\begin{equation}
Ei  =  -p \cdot \ln\bigg(\frac{1-p}{p}\bigg) \label{computationalModelEquation-i}
\end{equation}
Here, the term~$i$ is a placeholder for a parameter regulating~$E$. The remainder of this paper will discuss how a neural network can use this parameter for learning. Moreover, choosing~$i$ as the character for the parameter is not by chance. Section~\ref{Sec:ProcessModel} will consider this to be a complex number and will take advantage of the fact that~$i^2=-1$.

\section{Loss function}
\label{sec:LossFunction}
This section will develop a loss function based on the theoretical results derived above. Section~\ref{Sec:IntrinsicUncertainty} showed that for an angle of $\phi = \pi/4$, one of the two dual intertwined processes will know the quantity of the input, but not the sign. Typically, for a machine learning task, the goal is to match the magnitude of the teaching input with the output of a network. The sign of the teaching input is ignored, or tacitly assumed to be positive. As discussed above, there is no way of knowing both the magnitude and the sign of the teaching input. Therefore, assuming a binary teaching input~$t$, let the difference function $d(y,t)$ between~$t$ and network output~$y$ be defined as follows:
\begin{equation}
d(y,t) = (y-t+1) \cdot \frac{\pi}{4}
\label{differenceEquation}
\end{equation}
This function will output values in the interval $\big[0;\frac{\pi}{2}\big]$, and will be equal to one of the outer boundaries of the interval when the difference between network output and teaching input is maximum, with $|y-t|=1$. In case the network output is identical to the teaching input, $d(y,t)$ will be in the middle of the interval, $d(y,t)=\pi/4$. The angle defined by $d(y,t)$ can now be used as $\phi$, and its cosine as the observed input.

Next, resolving Eq.~\ref{computationalModelEquation} for the true probability~$p$ leads to the following sigmoidal expression for~$p$:
\begin{equation}
p = \frac{1}{1+\exp\left(-E/p\right)}
\label{sigmoidEquation}
\end{equation}
In this expression, the denominator~$p$ in the argument to the exponential function can be written as~$\sin(d)$, according to Eq.~\ref{pythagoreanEquation}. Furthermore, the energy~$E$ can be computed as the Kullback-Leibler divergence based on an input probability of~$\cos^2d$, using Eq.~\ref{KullbackLeiblerEquation2}. Here, the cosine value is squared to include the multiplication by two in Eq.~\ref{pythagoreanEquation}.
Performing these substitutions in Eq.~\ref{sigmoidEquation} produces the following sigmoid function:
\begin{equation}
p = \frac{1}{1+\exp\left(-D(\cos^2(d))/\sin(d)\right)}
\label{sigmoidEquationSinCos}
\end{equation}
With the difference between network output and teaching input defined according to Eq.~\ref{differenceEquation}, this function will output values in the interval $\left[\frac{1}{2}\,;1\right]$. When the network output is equal to the teaching input, the Kullback-Leibler divergence~$D$ in Eq.~\ref{sigmoidEquationSinCos} will be zero ($\cos^2(d)=0.5$), and the function will thus attain its minimum value of $0.5$. Therefore, the function defined by Eq.~\ref{sigmoidEquationSinCos} can be used as a loss function, which reaches its minimum of $0.5$ if, and only if, the Kullback-Leibler divergence is zero, assuming a radial distance measure between network output and teaching input, as defined in Eq.~\ref{differenceEquation}.

\section{Dual Process Model}
\label{Sec:ProcessModel}

\begin{figure*}[ht] % h-here, t-top, (b-bottom)
\vskip 0.2in
\begin{center}
%\centerline{\includegraphics[width=\columnwidth]{figures/DualProcessModel}}
\centerline{\includegraphics[width=\textwidth]{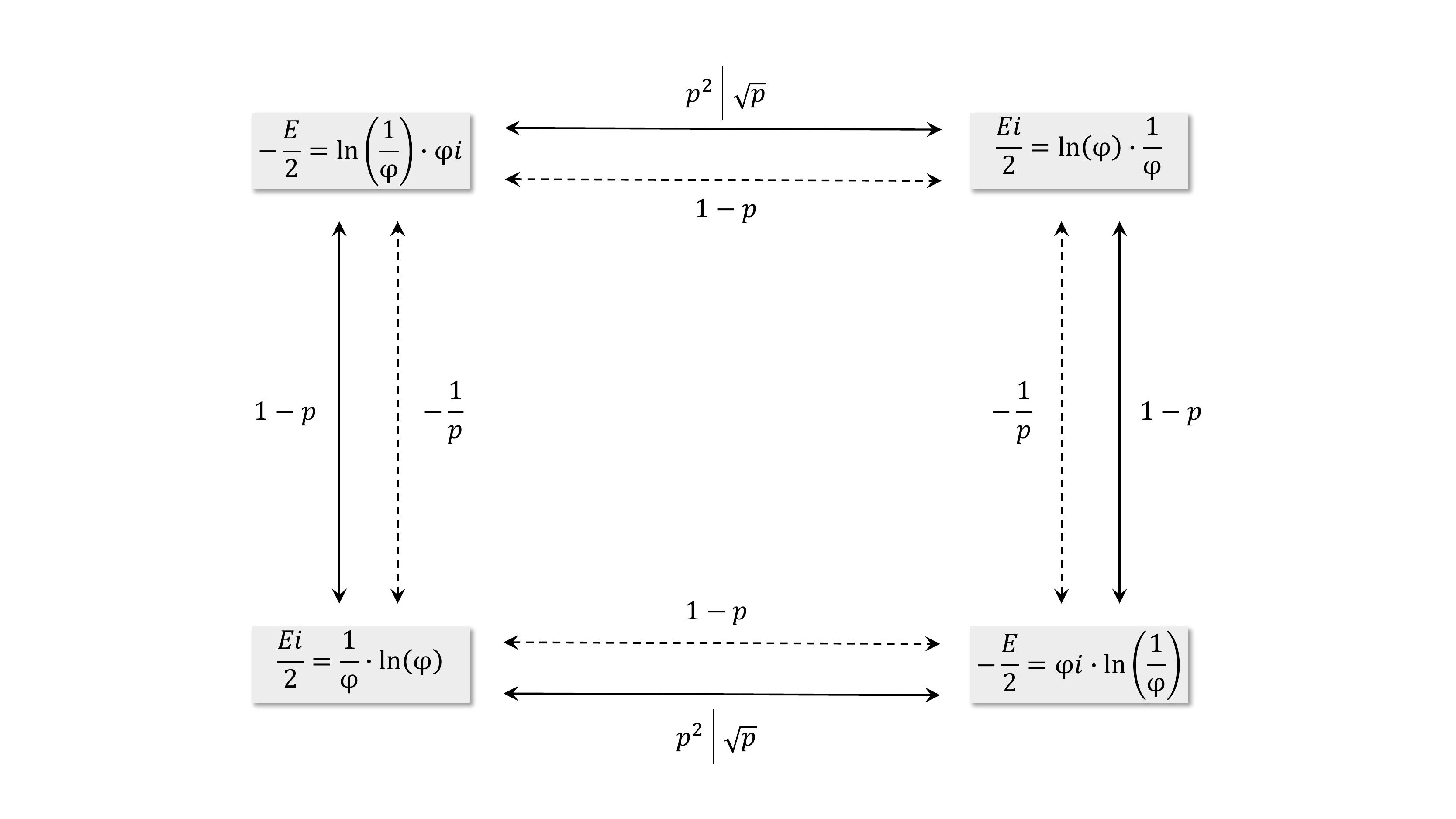}}
\caption{Equilibrium states and transition scheme of the dual process model, with the golden ratio $\varphi \approx 1.618$}
\label{dualProcessFigure}
\end{center}
\vskip -0.2in
\end{figure*}

Section~\ref{sec:LossFunction} showed that minimizing the Kullback-Leibler divergence towards~$p=0.5$, and thus maximizing the entropy, allows learning the magnitude of the teaching input. However, according to the theory set forth here, there is a dual process that maximizes the Kullback-Leibler divergence, and thus minimizes the entropy, to learn the sign of the teaching input. Both processes run in parallel, and are intertwined, with the output of one process being the input to the respective other process. They are complementary in the sense that their underlying true probabilities are complements, either~$p$ or $1-p$, and so are their measurements, as motivated in Section~\ref{Sec:MotivationApproach}. To illustrate this principle, Figure~\ref{dualProcessFigure} shows the transition scheme between different equilibrium states from the viewpoint of a process.
Each transition from one corner to another in the diagram in Figure~\ref{dualProcessFigure}, represents either a change in the underlying complementary, true probability, or a change in measurement, meaning measuring the complement instead. For example, the equation in the lower-left corner in Figure~\ref{dualProcessFigure} is related to Eq.~\ref{pythagoreanEquation}. This becomes clear when looking at the following equation:
\begin{equation}
\frac{Ei}{2} = -\frac{1}{\sqrt{2}} \cdot \ln\left(\frac{1}{\sqrt{2}}\right) \label{GoldenRatioPythagoreanEquation}
\end{equation}
Eq.~\ref{GoldenRatioPythagoreanEquation} is almost identical to Eq.~\ref{pythagoreanEquation}, with $\phi = \pi/4$, except that the complex term~$i$ has been added, similar to Eq.~\ref{computationalModelEquation-i}. Thus, Eq.~\ref{GoldenRatioPythagoreanEquation} describes an equilibrium state in which the amount of released energy equals the amount of absorbed energy. According to Section~\ref{Sec:IntrinsicUncertainty}, this is the state where the measured probability equals the true probability, which is the case for the golden ratio. Therefore, the term $1/\sqrt{2}$ needs to be multiplied by the constant $\sqrt{2} \cdot p_1$, where $p_1$ is the value of the golden ratio as defined in Eq.~\ref{p1Equation}, in order to obtain the golden ratio in a state of equilibrium. Expressing the minus sign with the inverse argument of the logarithm, in Eq.~\ref{GoldenRatioPythagoreanEquation}, and writing $p_1$ as $1/\varphi$, with $\varphi \approx 1.618$, produces the equation in the lower-left corner of Figure~\ref{dualProcessFigure}:
\begin{equation}
\frac{Ei}{2} = \frac{\ln(\varphi)}{\varphi} 
\label{GoldenRatioEquilibriumEquation}
\end{equation}

By going up vertically in the diagram in Figure~\ref{dualProcessFigure}, reaching the equation of the dual process in the upper-left corner, the original input, $\varphi$, becomes what can be considered the output of the dual process, and the original output, $1/\varphi$, is now the input, or measurement, of the dual process. The dual process is based on the complement of the true probability, $1-p$, thus the double arrow in Figure~\ref{dualProcessFigure}. Furthermore, the energy in the dual process flows in the opposite direction. Therefore, a multiplication by the complex number~$i$ introduces a minus sign for the energy and brings the complex number to the other side of the equation ($i^2=-1$). Moving further along to the upper-right corner of Figure~\ref{dualProcessFigure} by measuring the complement of the input, which involves squaring the input according to Eq.~\ref{complementPlusEquation}, returns to the energy computation of the original process. Therefore, a multiplication by~$i$ removes the minus sign and brings~$i$ back to the other side of the equation.

Taking the complement twice in a row, once for the true probability ($1-p$) and once for the measured input ($p^2$), is equivalent to looking at the same process from a different point of view. The next section will show how to exploit this fact for machine learning. The dashed lines in Figure~\ref{dualProcessFigure} show the corresponding state transitions for the opposing dual process, which uses the relation in Eq.~\ref{complementMinusEquation} to compute the complement of its input ($p=-1/p$).

\section{Regularization}
\label{Sec:Regularization}

A training method based on backpropagation adapts the network weights in a way that minimizes the loss, meaning the difference between network output and teaching input~\cite{lecun2012efficient}. Using gradient descent, training implies computing the gradient of a loss function~$L$, such as the loss given by Eq.~\ref{sigmoidEquationSinCos}, with respect to each network weight. A backpropagation method accomplishes this for one network layer at a time, iteratively, propagating the gradient back from the output layer to the input layer. To move along the gradient towards the minimum of the loss function, a delta is added to each weight, which has the following form, when adding also a momentum term:
\begin{equation}
\Delta w_{ij}(t) = -\eta \frac{\partial L}{\partial w_{ij}(t)} + \alpha \cdot \Delta w_{ij}(t-1)
\label{deltaWithMomentumEquation}
\end{equation}
In Eq.~\ref{deltaWithMomentumEquation}, $\Delta w_{ij}(t)$ denotes the delta added to each weight $w_{ij}$ between a node $i$ and a node $j$ in the network, at training iteration (or time)~$t$. The term $\partial L/\partial w_{ij}(t)$ is the partial derivative of the loss function with respect to $w_{ij}$, at time~$t$, which is multiplied with the learning rate~$\eta$. The sign of~$\Delta w_{ij}(t)$ is negative so that the loss function approaches its minimum. In practice, a momentum term describing the weight change at time~$t-1$, $\Delta w_{ij}(t-1)$, is commonly added. This term is typically multiplied by a weighting factor~$\alpha$, as seen in Eq.~\ref{deltaWithMomentumEquation}.
The general conception is that the momentum term improves stochastic gradient descent by dampening oscillations. However, according to the dual process model developed here, the actual reason for the performance improvement brought about by the momentum term lies in the gradient of the dual process, as discussed below.

As of yet, a conclusive theory for the optimal values of the learning rate~$\eta$ and the momentum weight~$\alpha$ has been lacking, although second order methods have been tried for example~\cite{bengio2012practical,sutskever2013importance}. Both parameters are often determined heuristically, either through empirical experiments or through search. Training results can be very sensitive to the value of the learning rate. For example, a small learning rate may produce a slow convergence, whereas a larger learning rate may result in the search passing over the minimum loss. Negotiating this delicate trade-off in the regularization of the training process can be time-consuming in practical applications. Literature seems to prefer learning rates around~$0.01$, although reported values differ by several orders of magnitude. For the momentum weight, higher values around $0.9$ are more common.

The proposed dual process model allows deriving theoretical values for both regularization parameters, learning rate~$\eta$ and momentum weight~$\alpha$. Eq.~\ref{GoldenRatioPythagoreanEquation} and Figure~\ref{dualProcessFigure} help to understand this. As explained in Section~\ref{Sec:ProcessModel}, the equilibrium equations in the lower-left and upper-right corner of Figure~\ref{dualProcessFigure}, represent the same process from different point of views. Bringing the factor $1/2$ from the left-hand side of Eq.~\ref{GoldenRatioPythagoreanEquation} to its right-hand side, by squaring the argument of the logarithm, it becomes evident that the measured probability is $1/2$ in the state of equilibrium. Therefore, this process minimizes the Kullback–Leibler divergence and maximizes the entropy. On the other hand, the dual process represented by the equilibrium equations in the upper-left and lower-right corner of Figure~\ref{dualProcessFigure}, does the opposite because it features a minus sign in front of both equations. It maximizes the Kullback–Leibler divergence and minimizes the entropy. A gradient in the dual process model is therefore a composite of the gradients of both processes, involving the gradient of one process and the negative gradient of its dual process. Each summand in the weight adjustment defined by Eq.~\ref{deltaWithMomentumEquation}, namely the partial derivative $\partial L/\partial w_{ij}(t)$ and the momentum term $\Delta w_{ij}(t-1)$, corresponds to a gradient of one of the dual processes.

The momentum weight~$\alpha$ follows from the results in Section~\ref{Sec:ProcessModel}. The multiplier for the equilibrium in Eq.~\ref{GoldenRatioPythagoreanEquation}, $1/\sqrt{2}$, is a gradient, when Eq.~\ref{GoldenRatioPythagoreanEquation} is considered a linear function with information input, $-\ln(x)$, and information output, $E/2$. The dual process has the same gradient, but with input and output reversed. Because both dual processes are intertwined, it is fair to assume that the dual process happens at time $t-1$, and that the current process at time~$t$ observes the output of its dual counterpart. Therefore, the multiplier in Eq.~\ref{GoldenRatioPythagoreanEquation} represents the gradient from the previous iteration. This gradient, and thus the delta at $t-1$, $\Delta w_{ij}(t-1)$, needs to be multiplied by a constant to obtain the golden ratio for the state of equilibrium, in Eq.~\ref{computationalModelEquation}, for which the measured probability equals the true probability. As in Section~\ref{Sec:ProcessModel}, this regularization can be computed as follows:
\begin{equation}
\alpha = \sqrt{2} \cdot p_1 \approx 0.874,
\label{momentumWeightEquation}
\end{equation}
where $p_1$ is again the value of the golden ratio in Eq.~\ref{p1Equation}, which provides the value for the momentum weight $\alpha \approx 0.874$.

The learning rate~$\eta$ can be derived from the momentum weight~$\alpha$ by converting the latter to the corresponding value of the dual process, following a processing chain in Figure~\ref{dualProcessFigure} of Section~\ref{Sec:ProcessModel}. For $\phi=\pi/2$, in Eq.~\ref{pythagoreanEquation}, $\sin(\phi)$ becomes one. Therefore, after multiplying with the momentum weight, the multiplier, or true probability, will be~$\alpha$. The true probability of the dual process is then given by the complement of~$\alpha$: $1-\alpha$. To make probabilities consistent with each other, the measured probability of the dual process needs to be squared according to Eq.~\ref{complementPlusEquation} in order to compute its complement. Taking the complement of the true probability and of the observed probability can be understood as looking at the same process from a dual point of view; see also Section~\ref{Sec:ProcessModel}. Consequently, applying these steps to the momentum weight~$\alpha$ results in the following expression for the learning rate~$\eta$:
\begin{equation}
\eta = (1-\alpha)^2 \approx 0.016
\label{learningRateEquation}
\end{equation}
This provides the value for the second regularization term, learning rate~$\eta$, with~$\eta \approx 0.016$.

\section{Discussion}

The previous section has shown how the regularization parameters of the delta learning rule given by Eq.~\ref{deltaWithMomentumEquation}, which are learning rate~$\eta$ and momentum weight~$\alpha$, can be derived from the dual process model proposed in Section~\ref{Sec:ProcessModel}. According to the theoretical framework set forth here, the delta rule combines the gradients of two dual processes. As already mentioned in~\cite{jaeger2020arXiv}, this goes beyond the traditional understanding according to which the momentum term produces a more stable gradient descent by smoothing weight changes over several iterations.
%In the theoretical framework presented in this paper, gradient descent is based on two dual processes.
While one process minimizes the Kullback–Leibler divergence ($p=0.5$) and maximizes the Shannon entropy ($p=0.5$), its dual counterpart does the opposite, by maximizing the Kullback–Leibler divergence and minimizing the entropy. This becomes evident when looking at Eq.~\ref{pythagoreanEquation} for which the equilibrium is reached when the observed probability inside the logarithm equals the true probability outside the logarithm, with $p=1/\sqrt{2}$. Squaring this value, meaning multiplying with the same value from the dual process for which observed and true probability switch their roles, leads to $p=0.5$.

The process that minimizes the Kullback–Leibler divergence ensures that the output equals the training input, while its dual counterpart that minimizes the Shannon entropy ensures that there is no uncertainty in the output. However, only both processes taken together can minimize the cross entropy. Each process alone has limitations. The process minimizing the Kullback–Leibler divergence may know that the output equals the training input, but it cannot know with absolute certainty whether the output should be zero or one. This is why the loss function given by Eq.~\ref{sigmoidEquationSinCos} is defined in such a way that it assumes its minimum for an angle of of~$45^\circ$. On the other hand, the process minimizing the Shannon entropy may know that the output is zero, but it cannot know if this is equal to the intended teaching input.

This reveals an inherent problem of the teaching input that has largely gone unnoticed so far in the literature. It is only possible to know either the teaching input signal, zero or one, or the actual teaching input, which could be identical to the teaching input signal or could equally well be its complement. It is only possible to know one or the other, but not both, which is reminiscent of Heisenberg's Uncertainty principle in physics~\cite{jaeger2013neurological}.

Under these theoretical considerations, the gradient adjustment by means of the delta learning rule, as defined by Eq.~\ref{deltaWithMomentumEquation}, becomes a composite of two gradient adjustments. On the one hand, the gradient is followed to minimize the Kullback–Leibler divergence. On the other hand, the reversed gradient of the dual process maximizing the Kullback–Leibler divergence is followed to minimize the entropy. This again explains the two different point of views, or directions for a process in Section~\ref{Sec:ProcessModel} and Section~\ref{Sec:Regularization}. After a successful training, both processes together have minimized the cross-entropy. However, their knowledge is distributed among them. While one process has learned to mimic the teaching input, the dual process has learned whether the teaching signal needs to be taken at face value or if it needs to be complemented.

The theoretical results in this paper confirm that cross-entropy is a profound loss function. However, rather than using cross-entropy directly as a loss function, it may be more appropriate to use it indirectly, via the sum of Kullback–Leibler divergence and Shannon entropy, following the dual process model laid out in Section~\ref{Sec:ProcessModel}. This can be achieved by using the loss function defined by Eq.~\ref{sigmoidEquationSinCos}, and applying the delta learning rule with momentum, as given by Eq.~\ref{deltaWithMomentumEquation}, with the specific values for learning rate~$\eta$ and momentum weight~$\alpha$ derived in Section~\ref{Sec:Regularization}.

\section{Conclusion}

This paper presents a theoretical analysis for the minimization of cross-entropy, which is one of the most popular loss functions used in combination with backpropagation for training artificial neural networks. The main result is a model comprising two dual processes, with one process minimizing the Kullback–Leibler divergence and the other process minimizing the Shannon entropy. The golden ratio as well as complex numbers play a major role in this model. Both are novel concepts in machine learning according to current knowledge. Moreover, specific values for learning rate and momentum weight follow from the model. The order of magnitude of their values is very similar to empirical values often used in practical experiments. Therefore, choosing these values for both regularization parameters improves the performance of gradient descent, and may help render an expensive hyper-parameter grid search redundant.

%%%%%%%%%%%%%%%%%%%%%%%%%%%%%%%%%%%%%%%%%%%%%%
%%%%%%%%%%%%%%%%%%%%%%%%%%%%%%%%%%%%%%%%%%%%%%

%\section*{Software and Data}

\section*{Acknowledgement}
This research was supported by the Intramural Research Program of the National Library of Medicine, National Institutes of Health. 

% Bibliography
\bibliography{StefanJaeger-DualProcess-arXiv2}
\bibliographystyle{icml2021}

\end{document}